\documentclass{article}

\usepackage{arxiv}

\usepackage[utf8]{inputenc} 
\usepackage[T1]{fontenc}    
\usepackage{hyperref}       
\usepackage{url}            
\usepackage{booktabs}       
\usepackage{amsfonts}       
\usepackage{nicefrac}       
\usepackage{microtype}      
\usepackage{lipsum}		
\usepackage{graphicx}
\usepackage{natbib}
\usepackage{doi}
\usepackage{amsmath}
\usepackage{float}

\title{The Power of Simplicity: Why Simple Linear Models Outperform Complex Machine Learning Techniques - Case of Breast Cancer Diagnosis}

\author{ \href{https://orcid.org/0009-0001-9793-9326}{\includegraphics[scale=0.06]{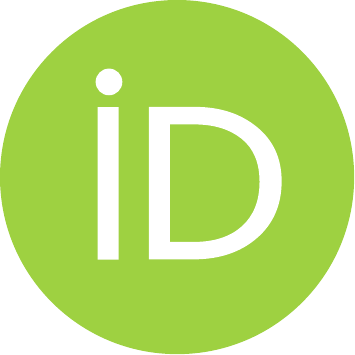}\hspace{1mm}Muhammad Arbab Arshad}\thanks{} \\
	Department of Computer Science\\
	Iowa State University\\
	Ames, IA 50010 \\
	\texttt{arbab@iastate.edu} \\
	\And
	{Sakib Shahriar} \\
	School of Computer Science\\
	University of Guelph\\
	Guelph, Ontario, Canada \\
	\texttt{shahrias@uoguelph.ca} \\
	\And
	{Khizar Anjum} \\
	Department of Electrical and Computer Engineering\\
	Rutgers University\\
	New Brunswick, NJ, USA \\
	\texttt{khizar.anjum@rutgers.edu} \\
}

\renewcommand{\shorttitle}{The power of simplicity}

\hypersetup{
pdftitle={A template for the arxiv style},
pdfsubject={q-bio.NC, q-bio.QM},
pdfauthor={David S.~Hippocampus, Elias D.~Striatum},
pdfkeywords={First keyword, Second keyword, More},
}

\begin{document}
\maketitle

\begin{abstract}
	This research paper investigates the effectiveness of simple linear models versus complex machine learning techniques in breast cancer diagnosis, emphasizing the importance of interpretability and computational efficiency in the medical domain. We focus on Logistic Regression (LR), Decision Trees (DT), and Support Vector Machines (SVM) and optimize their performance using the UCI Machine Learning Repository dataset. Our findings demonstrate that the simpler linear model, LR, outperforms the more complex DT and SVM techniques, with a test score mean of 97.28\%, a standard deviation of 1.62\%, and a computation time of 35.56 ms. In comparison, DT achieved a test score mean of 93.73\%, and SVM had a test score mean of 96.44\%. The superior performance of LR can be attributed to its simplicity and interpretability, which provide a clear understanding of the relationship between input features and the outcome. This is particularly valuable in the medical domain, where interpretability is crucial for decision-making. Moreover, the computational efficiency of LR offers advantages in terms of scalability and real-world applicability. The results of this study highlight the power of simplicity in the context of breast cancer diagnosis and suggest that simpler linear models like LR can be more effective, interpretable, and computationally efficient than their complex counterparts, making them a more suitable choice for medical applications.

\end{abstract}

\keywords{Algorithm Efficiency \and Model Interpretability \and Breast Cancer Diagnosis}

\section{Introduction}
Breast cancer is the most common cancer among women worldwide, accounting for approximately 25\% of all cancer cases \cite{bray2018global}. In 2020, there were an estimated 2.3 million new cases of breast cancer, with 685,000 deaths globally \cite{sung2021global}. Early detection and accurate diagnosis of breast cancer are crucial for improving patient outcomes, as the survival rates significantly increase when the disease is detected at an early stage \cite{cancer_facts_figures_2021}. Consequently, research into improving breast cancer diagnosis techniques has been a focal point in the medical community.

Traditionally, breast cancer diagnosis relies on a combination of clinical examination, mammography, ultrasound, and biopsy \cite{onega2014breast}. Mammography, which is an X-ray imaging technique, is the primary screening tool for breast cancer detection. While mammography has been shown to reduce breast cancer mortality, it has some limitations, such as a higher rate of false positives and false negatives, particularly in women with dense breast tissue \cite{sprague2014prevalence}. These limitations can lead to unnecessary biopsies, additional tests, and psychological distress for patients \cite{nelson2016harms}.

In recent years, there has been a growing interest in developing and refining computational techniques to aid in breast cancer diagnosis. Machine learning, a subset of artificial intelligence, has shown great promise in improving the accuracy and efficiency of breast cancer detection and classification \cite{bejnordi2017diagnostic}. However, the adoption of complex machine learning algorithms sometimes introduces issues such as overfitting, lack of interpretability, and increased computational requirements, which may hinder their practical application \cite{choi2020comparing}.

This study examines the power of simplicity in breast cancer diagnosis by comparing the performance of simple linear models, such as logistic regression, with more complex machine learning techniques, including support vector machines (SVM) and decision trees. By demonstrating the efficacy of simple linear models in breast cancer diagnosis, we aim to encourage the development of interpretable, computationally efficient, and easily implementable diagnostic tools that can aid healthcare professionals in making more accurate diagnoses and improving patient outcomes.

In recent studies, simple linear models have been found to achieve comparable or even superior performance in various medical diagnostic tasks when compared to more complex machine learning techniques \cite{boateng2019review, ma2021comparison}. This can be attributed to several factors, including the avoidance of overfitting, the simplicity of the model structure, and the ease of interpretation of the results \cite{choi2020comparing}. Moreover, linear models can handle collinearity and multicollinearity issues more effectively, which are common in medical datasets \cite{gurcan2015histopathological}. These advantages make simple linear models particularly attractive for clinical applications, where interpretability and generalizability are crucial for medical decision-making.

Furthermore, the computational efficiency of simple linear models allows for faster training and prediction times, making them suitable for real-time clinical applications and large-scale data analysis. The reduced complexity of these models also facilitates their integration into existing clinical workflows, as they require less specialized knowledge for implementation and maintenance \cite{ma2021comparison}. In this context, the power of simplicity becomes evident, as it enables the development of practical, interpretable, and efficient diagnostic tools that can better serve healthcare professionals and patients.

In this study, we will demonstrate the effectiveness of simple linear models in the context of breast cancer diagnosis using a dataset from the UCI Machine Learning Repository. Our results will highlight the potential of these models to outperform more complex machine learning techniques and provide valuable insights into the benefits of simplicity in medical diagnosis.

The Breast Cancer Wisconsin (Diagnostic) Data Set (WDBC) is a widely used benchmark dataset in the field of machine learning for the development and evaluation of breast cancer classification models \cite{frank2010uci}. The dataset was created by Dr. William H. Wolberg, Dr. W. Nick Street, and Dr. Olvi L. Mangasarian at the University of Wisconsin-Madison and was first made publicly available in 1995 through the UCI Machine Learning Repository \cite{street1993nuclear}.

The WDBC dataset contains 569 instances, each representing a separate breast cancer case. For each case, there are 32 attributes, including an ID number, a diagnosis (either malignant or benign), and 9 real-valued features derived from digitized fine needle aspirate (FNA) images of breast masses \cite{wolberg1994machine}. The FNA procedure involves using a thin needle to collect cell samples from a breast mass, which can then be examined under a microscope to determine the presence of cancer \cite{pisano2001fine}. The 9 features in the dataset are computed from these FNA images, providing insights into the morphological characteristics of cell nuclei, such as texture, smoothness, compactness, symmetry, and fractal dimension \cite{haralick1973textural}. These features are divided into three groups: mean, standard error, and worst (mean of the three largest values) for each of the ten primary feature types \cite{cruz2006applications}.

The Breast Cancer Wisconsin (Diagnostic) Data Set has been extensively used for the development, evaluation, and comparison of various machine learning algorithms in the context of breast cancer diagnosis. Researchers have employed techniques such as logistic regression, support vector machines, decision trees, neural networks, k-nearest neighbors, and ensemble methods to create classification models that predict the malignancy or benignity of breast masses based on the provided features \cite{abdel2016breast} \cite{chaurasia2017novel} \cite{kourou2015machine}. These models have demonstrated varying degrees of success, with some achieving accuracy rates of over 95

In this study, we aim to demonstrate the power of simplicity in breast cancer diagnosis by focusing on the performance of a simple linear model, logistic regression, and comparing it with more complex machine learning techniques, such as support vector machines and decision trees. By analyzing the WDBC dataset, we hope to showcase the efficacy of simple linear models in accurately predicting the malignancy or benignity of breast masses and provide insights into the benefits of simplicity in medical diagnosis. These insights can contribute to the development of more interpretable, computationally efficient, and easily implementable diagnostic tools, ultimately leading to better patient outcomes.



\section{Data Preprocessing and Exploration}

\subsection{Data Cleanup Process}

The raw data obtained from the Breast Cancer Wisconsin (Diagnostic) Data Set required some preprocessing to ensure optimal performance of the machine learning models. The data cleanup process involved the following steps:

\begin{enumerate}
\item \textbf{Balancing the dataset:} To prevent any bias towards a specific class, the dataset was balanced to have an equal number of benign and malignant instances. This was achieved by either oversampling the minority class, undersampling the majority class, or using a combination of both techniques.

\item \textbf{Dropping rows with missing values:} The 'Bare Nuclei' column had 16 instances with missing values. Since this number is relatively small compared to the total number of instances, these rows were dropped from the dataset.

\item \textbf{Converting 'Class' values to binary:} The 'Class' column originally had values of 2 for benign and 4 for malignant instances. These values were converted to binary, where 0 represents benign and 1 represents malignant cases.

\item \textbf{Updating data types:} The data types for the 'Bare Nuclei' and 'Class' columns were updated to ensure consistency and facilitate further analysis. 

\item \textbf{Removing 'Id' column:} The 'Id' column, which contains unique identifiers for each instance, was removed from the dataset, as it does not provide any useful information for the analysis.
\end{enumerate}

\subsection{Data Exploration and Visualization}

In order to attain a comprehensive understanding of the dataset and its inherent characteristics, a variety of visualizations were generated. These visualizations serve to elucidate the distribution of each feature, as well as the relationships between features for both benign and malignant instances.

\subsubsection{Parallel Coordinate Plot}

\begin{figure}[ht]
\centering
\includegraphics[width=\linewidth]{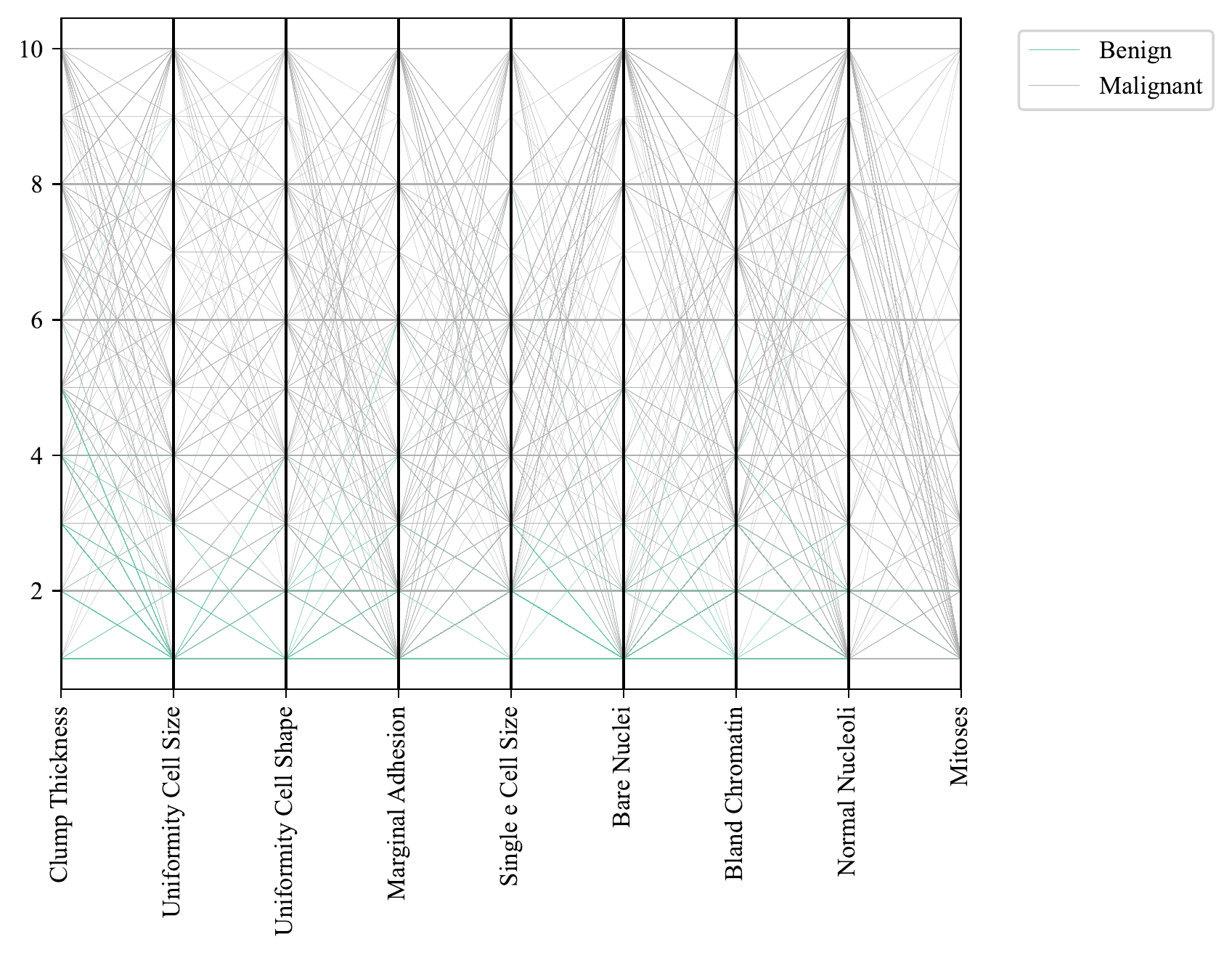}
\caption{Parallel coordinate plot of the Breast Cancer Wisconsin (Diagnostic) Data Set.}
\label{fig:parallel_coordinate_plot}
\end{figure}

A parallel coordinate plot was employed to visualize the distribution of each feature for both classes, as suggested by \cite{inselberg2008parallel}. In this plot, each row in the data table is represented as a line, with the axes corresponding to the features. This graphical method can be utilized to identify any patterns or trends in the data that may prove advantageous for classification purposes. The visualization is presented in Figure \ref{fig:parallel_coordinate_plot}. This visual representation of the data allows for the identification of any patterns or trends that could be beneficial for classification. It was observed that benign samples are more likely to exhibit lower values for the features, whereas malignant samples are more likely to demonstrate higher values. Consequently, the features could serve to distinguish between the two classes. This analysis provides a high-level overview of the dataset, and a more in-depth exploration of the data to uncover additional patterns will be conducted in the subsequent section.

The parallel coordinate plot in Figure \ref{fig:parallel_coordinate_plot} provides a comprehensive view of the distribution of features for both benign and malignant instances. As our study focuses on the power of simplicity in breast cancer diagnosis, it is important to identify patterns and trends in the data that can be effectively captured by simple linear models like logistic regression. Observations from the parallel coordinate plot indicate that there are distinctions between the feature values for benign and malignant samples. This suggests that simple linear models could potentially capture these relationships and provide accurate classification results, highlighting the power of simplicity in medical diagnosis. In the following sections, we will further explore the data and build our models, comparing the performance of logistic regression with more complex machine learning techniques, such as support vector machines and decision trees.

\subsubsection{Distribution of Each Feature for Both Classes}

Another visualization generated was the distribution of each feature for both benign and malignant instances. This type of plot provides an overview of the data's characteristics \cite{stigler1986history}, helping to identify any significant differences between the two classes for each feature.

In this analysis, we will discuss the distribution of the dataset's features and their potential impact on the prediction of breast cancer using simple linear models like logistic regression. The dataset consists of 239 benign and 239 malignant samples, with each sample described by nine features: Clump Thickness, Uniformity Cell Size, Uniformity Cell Shape, Marginal Adhesion, Single Epithelial Cell Size, Bare Nuclei, Bland Chromatin, Normal Nucleoli, and Mitoses.

\begin{itemize}
\item \textbf{Clump Thickness:} The mean clump thickness for benign samples is 2.95, while for malignant samples, it is 7.19. The higher mean for malignant samples indicates that the clump thickness is generally greater for malignant tumors. The standard deviation for benign and malignant samples is 1.59 and 2.44, respectively, indicating that malignant samples have a more varied clump thickness.

\item \textbf{Uniformity Cell Size:} The mean for benign samples is 1.29, while for malignant samples, it is 6.58. Malignant samples have a higher mean, indicating a greater variation in cell size. The standard deviation is 0.82 for benign samples and 2.72 for malignant samples, showing that malignant samples exhibit more variation in cell size.

\item \textbf{Uniformity Cell Shape:} The mean for benign samples is 1.44, while for malignant samples, it is 6.56. This shows that malignant samples have more varied cell shapes. The standard deviation is 0.93 for benign samples and 2.57 for malignant samples, again indicating a more significant variation in cell shape for malignant samples.

\item \textbf{Marginal Adhesion:} The mean for benign samples is 1.34, while for malignant samples, it is 5.59. Malignant samples have a higher mean, suggesting a greater degree of adhesion in malignant tumors. The standard deviation is 0.89 for benign samples and 3.20 for malignant samples, indicating a more considerable variation in adhesion for malignant samples.

\item \textbf{Single Epithelial Cell Size:} The mean for benign samples is 2.13, while for malignant samples, it is 5.33. Malignant samples have a higher mean, indicating a larger cell size. The standard deviation is 1.00 for benign samples and 2.44 for malignant samples, showing a larger variation in cell size for malignant samples.

\item \textbf{Bare Nuclei:} The mean for benign samples is 1.32, while for malignant samples, it is 7.63. This indicates a greater presence of bare nuclei in malignant samples. The standard deviation is 1.16 for benign samples and 3.12 for malignant samples, suggesting more variation in the number of bare nuclei in malignant samples.

\item \textbf{Bland Chromatin:} The mean for benign samples is 2.05, while for malignant samples, it is 5.97. Malignant samples have a higher mean, indicating a more significant presence of bland chromatin. The standard deviation is 0.99 for benign samples and 2.28 for malignant samples, indicating a larger variation in the amount of bland chromatin in malignant samples.

\item \textbf{Normal Nucleoli:} The mean for benign samples is 1.23, while for malignant samples, it is 5.86. This shows that malignant samples tend to have more normal nucleoli. The standard deviation is 0.87 for benign samples and 3.35 for malignant samples, indicating a more significant variation in the presence of normal nucleoli in malignant samples.

\item \textbf{Mitoses:} The mean for benign samples is 1.04, while for malignant samples, it is 2.60. This indicates that malignant samples tend to have a higher rate of mitosis. The standard deviation is 0.31 for benign samples and 2.56 for malignant samples, showing a more significant variation in mitosis rates for malignant samples.
\end{itemize}

\begin{figure}[ht]
\centering
\includegraphics[width=\linewidth]{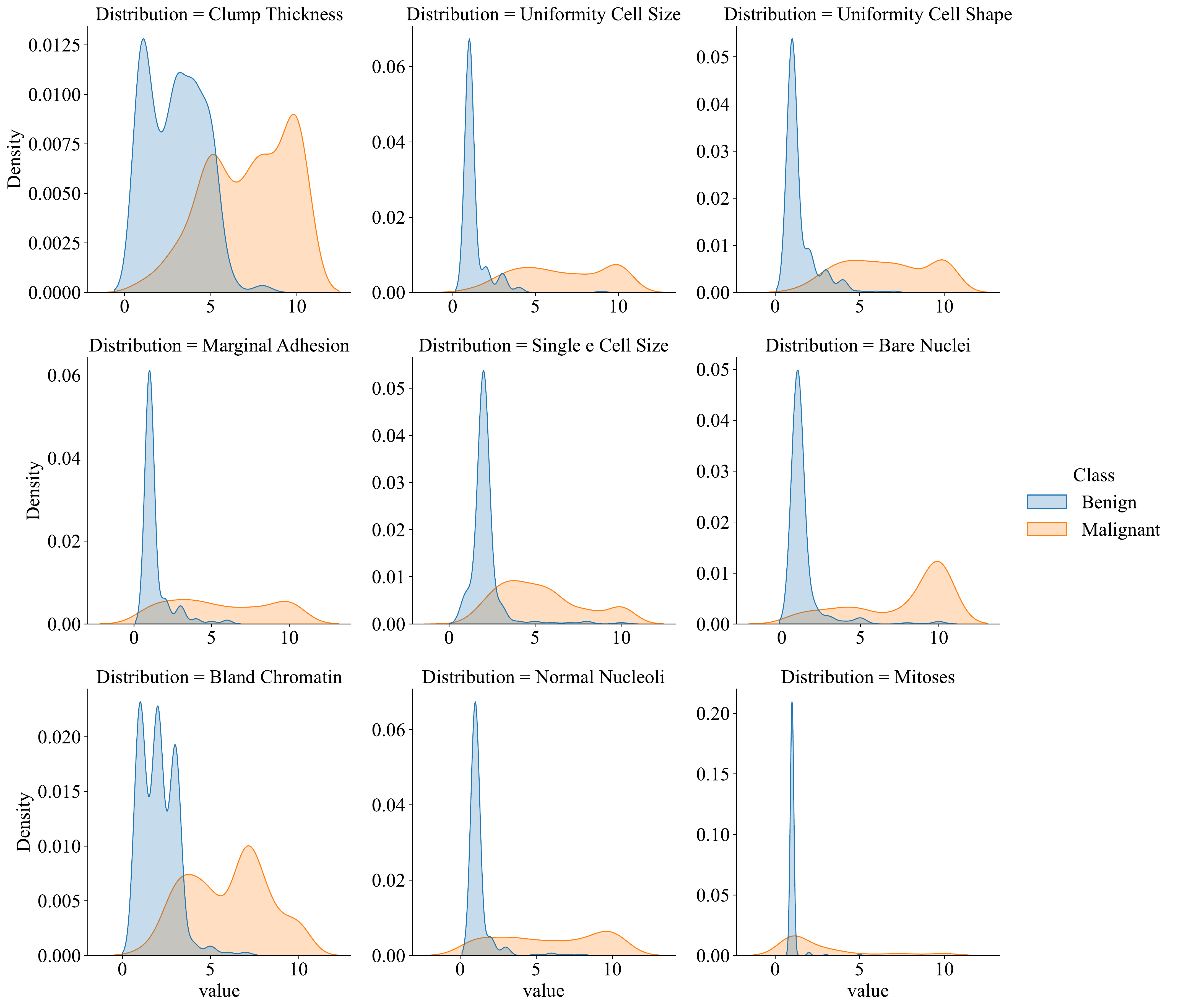}
\caption{Distribution of each feature for both benign and malignant instances.}
\label{fig:feature_distribution}
\end{figure}

In conclusion, the analysis of the Wisconsin Breast Cancer dataset shows that most features have a higher mean and larger standard deviation for malignant samples compared to benign samples. The visualization is given in Figure \ref{fig:feature_distribution}.
These differences in the distribution of the features indicate that they could be useful in distinguishing between benign and malignant tumors when applying simple linear models like logistic regression. By training the models with these features, we can potentially harness the power of simplicity to improve the accuracy and reliability of breast cancer detection \cite{han2022data}.

These visualizations provide valuable insights into the dataset's structure and can inform the selection of appropriate machine learning models and feature engineering techniques for classification. Our study aims to demonstrate the effectiveness of simple linear models like logistic regression in accurately diagnosing breast cancer based on these feature distributions.

\section{Machine Learning Models and Experimental Setup}

In this section, we will discuss the machine learning models used in this analysis and the experimental setup for evaluating the models' performance.     

\subsection{Logistic Regression}

Logistic Regression (LR) is a linear model for binary classification, which estimates the probability of an instance belonging to a specific class \cite{hosmer2013applied}. Given a set of input features $\textbf{x} = (x_1, x_2, ..., x_n)$, logistic regression computes the probability of an instance belonging to the positive class as follows:

\begin{equation}
P(y = 1|\textbf{x}) = \frac{1}{1 + e^{-(\beta_0 + \beta_1 x_1 + \beta_2 x_2 + ... + \beta_n x_n)}}
\end{equation}

where $\beta_0, \beta_1, ..., \beta_n$ are the model parameters that are learned from the training data. The logistic function, also known as the sigmoid function, maps the linear combination of input features to a probability value between 0 and 1. The decision boundary for logistic regression is linear in the feature space.

To train the logistic regression model, we optimize the model parameters using maximum likelihood estimation, which aims to maximize the likelihood of the observed data given the model parameters \cite{bishop2006pattern}. Regularization techniques, such as L1 or L2 regularization, can be applied to prevent overfitting and improve generalization \cite{tibshirani1996regression}.

\subsection{Support Vector Machines}

Support Vector Machines (SVM) is a powerful technique for binary classification, which aims to find the optimal separating hyperplane that maximizes the margin between the two classes \cite{cortes1995support}. Given a set of training data $(\textbf{x}_i, y_i)$, where $\textbf{x}_i$ is the feature vector and $y_i \in {-1, 1}$ is the corresponding class label, the primal optimization problem for SVM can be formulated as:

\begin{equation}
\min_{\textbf{w}, b} \frac{1}{2} ||\textbf{w}||^2 + C \sum_{i=1}^N \xi_i
\end{equation}
\begin{equation}
\text{subject to} \quad y_i(\textbf{w}^T \textbf{x}_i + b) \geq 1 - \xi_i, \quad \xi_i \geq 0
\end{equation}

where $\textbf{w}$ is the weight vector, $b$ is the bias term, $\xi_i$ is the slack variable for each instance, and $C$ is a regularization parameter that controls the trade-off between maximizing the margin and minimizing classification errors. The optimization problem is typically solved using the dual formulation, which involves only the dot products of the input features \cite{scholkopf1999advances}.

For non-linearly separable data, SVM can use kernel functions to implicitly map the input data into a higher-dimensional feature space, where a linear decision boundary can be found \cite{shawe2004kernel}. Some common kernel functions include the linear, polynomial, radial basis function (RBF), and sigmoid kernels.

\subsection{Decision Trees}

Decision Trees (DT) are a popular machine learning technique for classification and regression tasks, which recursively split the input feature space into regions based on feature values and assign a class label to each region \cite{quinlan1986induction}. The decision tree is represented as a tree structure, where internal nodes correspond to feature tests, and leaf nodes correspond to class labels.

The construction of a decision tree is a top-down process, where the most discriminative feature is chosen at each step to split the data into subsets. The selection of the best feature to split the data is based on a splitting criterion, such as Gini impurity or information gain, which measures the homogeneity of the resulting subsets \cite{breiman1984classification}. The process is repeated recursively for each subset until a stopping criterion is met, such as reaching a maximum tree depth, a minimum number of instances in the leaf, or no further improvement in the splitting criterion.

\begin{equation}
\text{Gini Impurity} = 1 - \sum_{i=1}^C p_i^2
\end{equation}

\begin{equation}
\text{Information Gain} = H(\textbf{X}) - \sum_{v \in \textbf{V}} \frac{|X_v|}{|X|} H(X_v)
\end{equation}

where $C$ is the number of classes, $p_i$ is the proportion of instances of class $i$ in a subset, $H(\textbf{X})$ is the entropy of the data set $\textbf{X}$, $\textbf{V}$ is the set of possible values for the feature, and $X_v$ is the subset of instances with the feature value $v$.

Decision trees can be prone to overfitting, especially when the tree is deep and complex. To mitigate overfitting, pruning techniques can be applied, which involve removing branches of the tree that provide little improvement in the splitting criterion \cite{chen2009pruning}. Ensemble methods, such as Random Forests and Gradient Boosting, can also be used to improve the performance and robustness of decision trees by combining multiple trees into a single model \cite{friedman2001greedy} \cite{breiman2001random}

\section{Methodology}

The methodology used in this study comprises several steps, including hyperparameter identification, grid search, and cross-validation. This process was applied to each of the three machine learning models - Logistic Regression, Support Vector Machines, and Decision Trees - to optimize their performance and enable a fair comparison.

\subsection{Hyperparameter Identification}

The first step in the methodology was to identify the key hyperparameters for each machine learning model. Hyperparameters are external configurations that cannot be learned by the model during training, and they significantly influence the performance of the model. Identifying the most relevant hyperparameters for each model is crucial for successful optimization.

\subsection{Grid Search}

Grid search is a technique used to perform an exhaustive search of the hyperparameter space to find the best combination of hyperparameter values for a given model. The process involves defining a range of values for each hyperparameter and evaluating the model's performance with each combination. The best set of hyperparameters is the one that results in the highest performance metric, such as accuracy or F1 score.

\subsubsection{Support Vector Machines (SVM)}

For the SVM model, the following hyperparameters were identified:

\begin{itemize}
\item \textbf{C:} The regularization parameter, which controls the trade-off between maximizing the margin and minimizing classification errors. The range of values considered for C was [0.1, 0.5, 1, 3, 9, 100].

\item \textbf{gamma:} The kernel coefficient for non-linear kernels, which controls the shape of the decision boundary. The range of values considered for gamma was [0.1, 1, 10].

\item \textbf{kernel:} The function used to transform the input data into a higher-dimensional space. The possible kernels considered were ['linear', 'poly', 'rbf', 'sigmoid'].
\begin{figure}[ht]
	\centering
	\includegraphics[width=\textwidth]{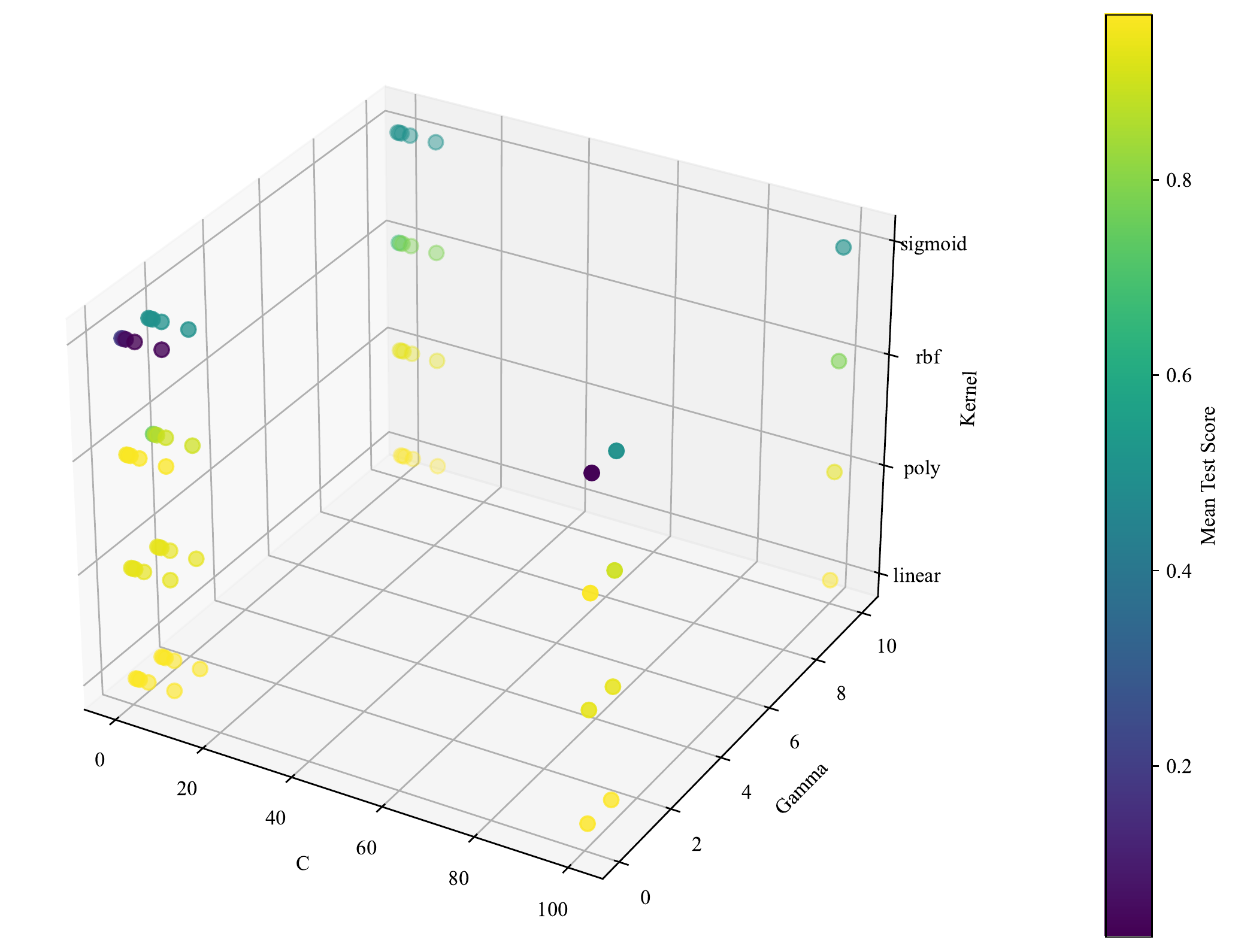}
	\caption{Grid search results for SVM}
	\label{fig:svm_grid_search}
	\end{figure}

\end{itemize}

The grid search identified the best hyperparameters for the SVM model as:

\begin{itemize}
\item C: 9
\item gamma: 0.1
\item kernel: linear
\end{itemize}

\subsubsection{Decision Trees}

For the Decision Trees model, the following hyperparameters were identified:

\begin{itemize}
\item \textbf{Max Depth:} The maximum depth of the tree, which controls the complexity of the model and prevents overfitting. The range of values considered for Max Depth was [1, 2, 5, 10, 15, 20, 30, 50, 100].

\item \textbf{Min Samples Split:} The minimum number of samples required to split an internal node. The range of values considered for Min Samples Split was [2, 5, 10, 15, 20, 30, 50].

\item \textbf{Min Samples Leaf:} The minimum number of samples required to be at a leaf node. The range of values considered for Min Samples Leaf was [1, 2, 5, 10, 15, 20,30, 50].
\end{itemize}

\begin{figure}[ht]
	\centering
	\includegraphics[width=\textwidth]{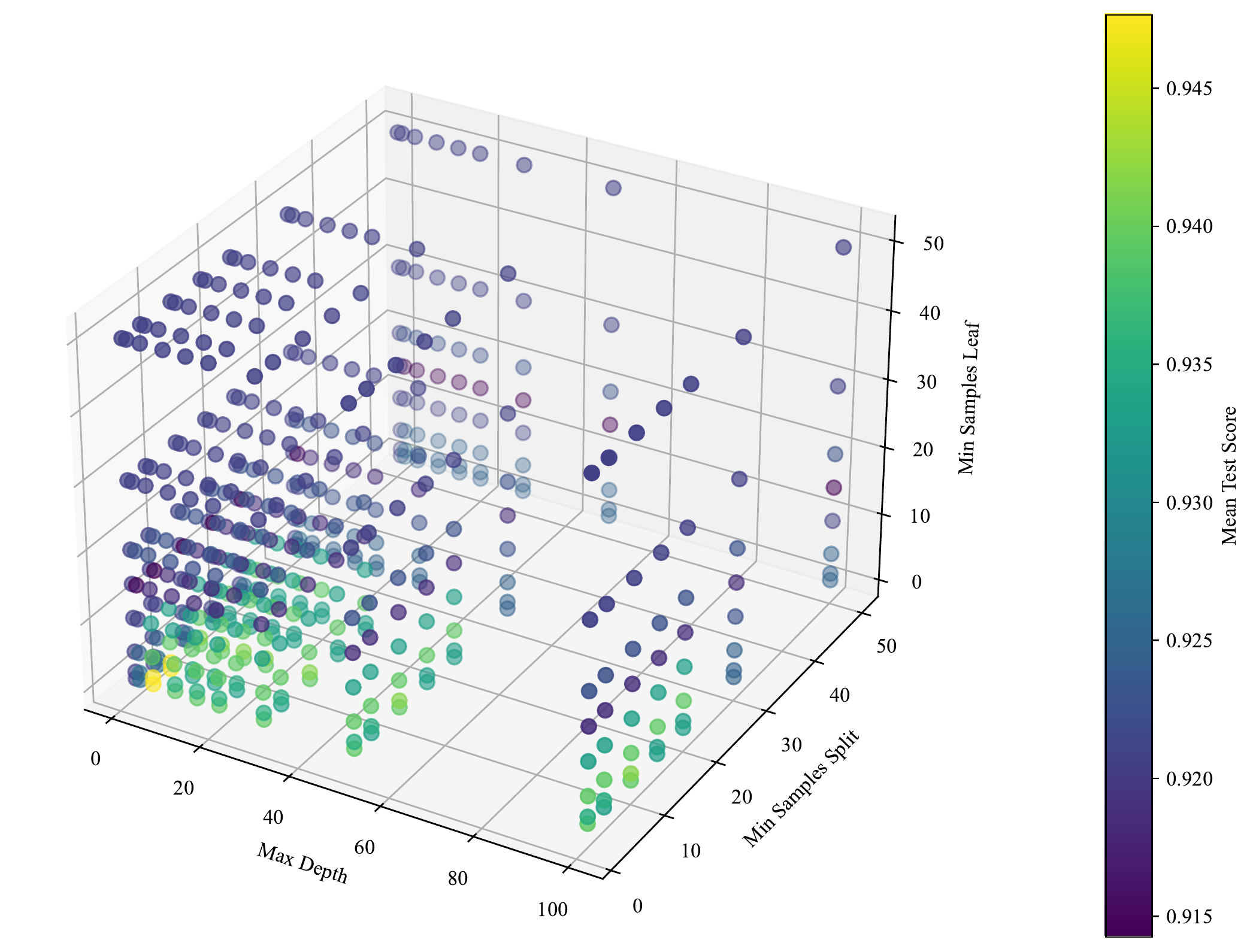}
	\caption{Grid search results for Decision Trees}
	\label{fig:dt_grid_search}
	\end{figure}

The grid search identified the best hyperparameters for the Decision Trees model as:

\begin{itemize}
\item Max Depth: 5
\item Min Samples Split: 2
\item Min Samples Leaf: 2
\end{itemize}

\subsubsection{Logistic Regression}

For the Logistic Regression model, the following hyperparameters were identified:

\begin{itemize}
\item \textbf{C:} The inverse of regularization strength, with smaller values specifying stronger regularization. The range of values considered for C was [0 … 10], incremented by 0.5.

\item \textbf{Solver:} The algorithm used for optimization. The possible solvers considered were ['lbfgs', 'newton-cg', 'newton-cholesky', 'sag', 'saga'].

\item \textbf{Penalty:} The norm used in the penalization. The possible penalties considered were ['none', 'l1', 'l2'].

\end{itemize}

\begin{figure}[ht]
	\centering
	\includegraphics[width=\textwidth]{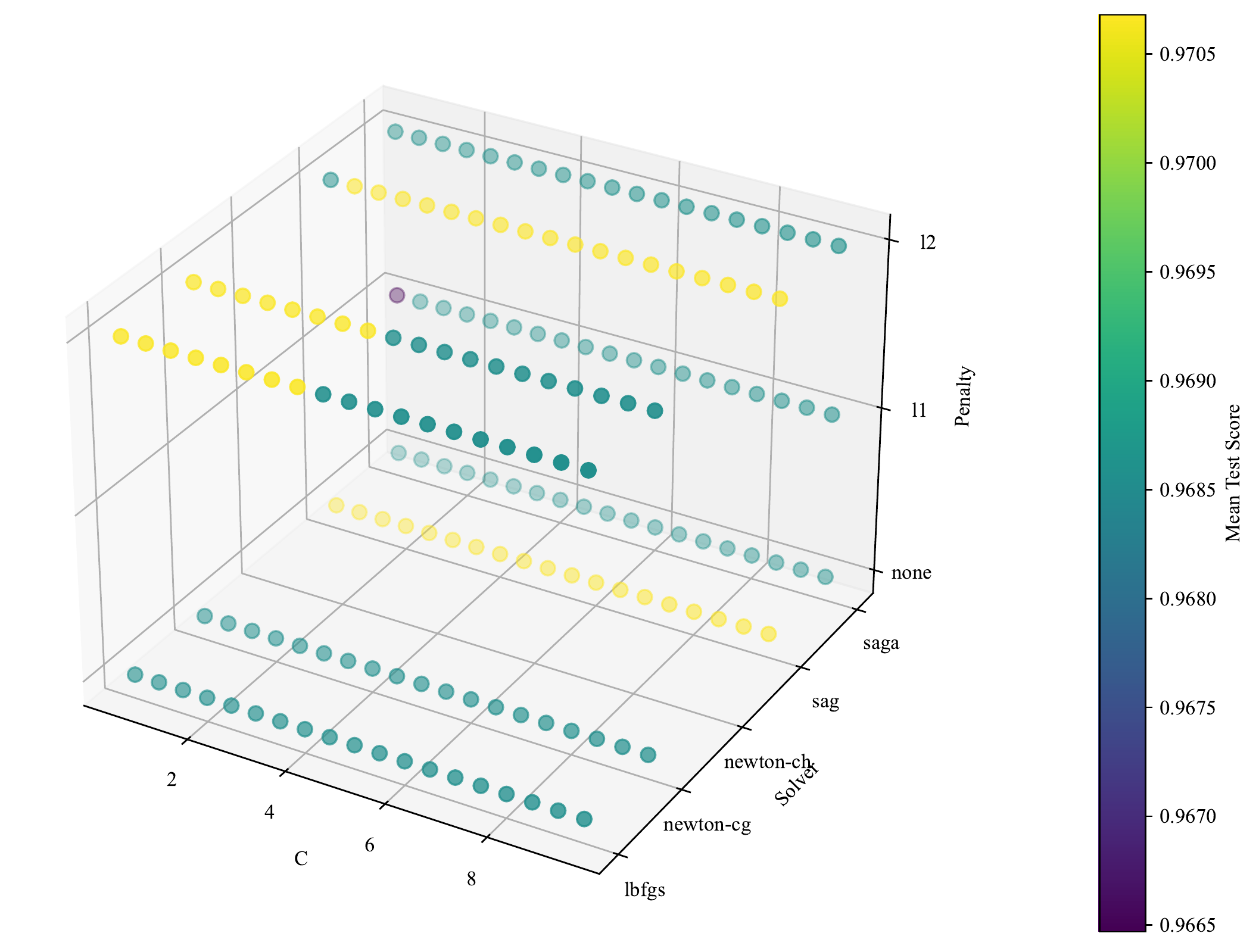}
	\caption{Grid search results for Logistic Regression}
	\label{fig:lr_grid_search}
	\end{figure}

The grid search identified the best hyperparameters for the Logistic Regression model as:

\begin{itemize}
\item C: 0.5
\item Solver: sag
\item Penalty: none
\end{itemize}

\subsection{Cross-Validation}

To assess the performance of the optimized models, we employed 10-fold cross-validation. This technique involves dividing the dataset into ten equal parts or folds, training the model on nine of these folds, and testing the model on the remaining fold. This process is repeated ten times, with each fold used as the test set once. The mean and standard deviation of the performance metric across the ten iterations are used to evaluate the model's performance. Cross-validation helps to reduce overfitting and provides a more reliable estimate of the model's performance on unseen data.

\section{Model Comparison and Analysis}

\begin{figure}[ht]
	\centering
	\includegraphics[width=\textwidth]{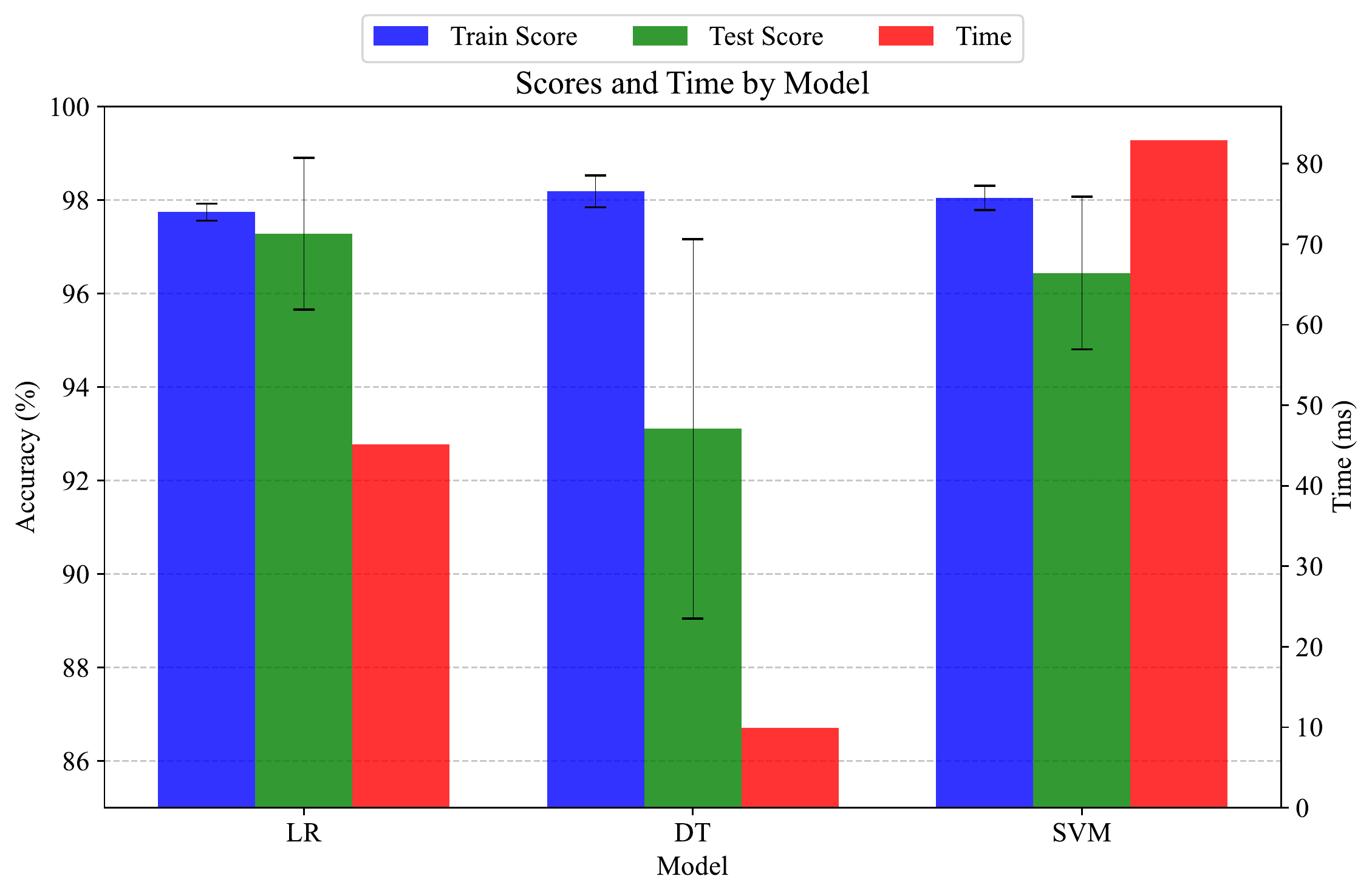}
	\caption{Model performance comparison}
	\label{fig:model_comparison}
	\end{figure}

The performance of the three models, Logistic Regression (LR), Decision Trees (DT), and Support Vector Machines (SVM), is illustrated in Figure~\ref{fig:model_comparison}. The figure presents the train and test scores, standard deviations, and computation time for each model. Based on these metrics, we can compare the advantages and disadvantages of each model and analyze why Logistic Regression performed the best. A summary table is given in Table~\ref{table:model_comparison}.

\subsection{Performance Comparison}

\begin{itemize}
\item \textbf{Logistic Regression (LR):} LR demonstrates the highest performance, with a test score mean of 97.28\% and the lowest test score standard deviation of 1.62\%. The computation time for LR is 35.56 ms. With superior test score mean and standard deviation, LR is especially suitable for binary classification problems. Although its computation time is moderate compared to other models, LR has a less complex hyperparameter tuning process than Support Vector Machines.

\item \textbf{Decision Trees (DT):} Despite having the highest train score mean (98.19\%), DT's test score mean is the lowest among the three models (93.73\%). The test score standard deviation is 3.08\%, and the computation time for DT is the least expensive at 8.97 ms. DT exhibits a lower test score mean and a tendency for overfitting, as indicated by the discrepancy between train and test scores. Nonetheless, DT models are suitable for both numerical and categorical data and require the least computation time among the three models. The complexity of tuning Decision Trees is moderate.

\item \textbf{Support Vector Machines (SVM):} SVM exhibits consistent performance, with a test score mean of 96.44\% and a test score standard deviation of 1.63\%. However, it has a longer computation time of 82.33 ms compared to LR and DT. SVM models are effective in handling non-linear relationships and managing high-dimensional spaces with appropriate kernel functions. Nevertheless, they require the slowest computation time among the three models and have a more complex hyperparameter tuning process, necessitating careful parameter selection.
\end{itemize}
	
\section{The Power of Simplicity: Linear Models Outperforming Complex Techniques}

Our analysis above demonstrates that Logistic Regression (LR) outperforms both Decision Trees (DT) and Support Vector Machines (SVM) in the context of breast cancer diagnosis. This section delves deeper into the reasons behind LR's superior performance and the implications of using simpler models for classification tasks.

\subsection{Interpretability and Complexity}

One key advantage of LR over DT and SVM is its simplicity and interpretability. LR models are easy to understand, as the coefficients of the model directly indicate the effect of each feature on the outcome. This allows for better clinical decision-making, as the model's predictions can be directly traced back to the input features.

In contrast, DT and SVM models involve more complex decision boundaries and are harder to interpret. The complexity of these models can lead to overfitting, as seen in DT's discrepancy between train and test scores. The higher complexity of SVM also results in a longer computation time and more challenging hyperparameter tuning.

\subsection{Generalization and Robustness}

LR's linear decision boundary allows for better generalization, as demonstrated by its higher test score mean and lower standard deviation. By avoiding overfitting, LR models are more robust to variations in the data and can better handle new, unseen samples. This is crucial in the context of medical diagnosis, where models must be able to handle a wide range of patients and conditions.

In contrast, the more complex decision boundaries of DT and SVM may lead to overfitting and reduced generalization capabilities. The discrepancy between train and test scores for DT indicates the model's inability to generalize well to new data. SVM, while having consistent performance, still falls short of LR in terms of test score mean and has a longer computation time.

\subsection{Scalability and Computational Efficiency}

The computational efficiency of LR is another advantage over more complex models like SVM. Although DT has the least computation time among the three models, it suffers from lower test score mean and overfitting issues. LR, on the other hand, balances computational efficiency and performance, making it an ideal choice for large-scale applications and real-world settings.

SVM models, while effective at handling non-linear relationships and high-dimensional spaces, have the slowest computation time among the three models. This can be a significant drawback, particularly when dealing with large datasets or when rapid diagnosis is critical.

\subsection{Tradeoffs and Model Selection}
On the other hand, Decision Trees and Support Vector Machines showed certain advantages, such as faster computation times and the ability to handle non-linear relationships. However, these benefits were offset by other disadvantages, such as overfitting in the case of Decision Trees and slower computation times for Support Vector Machines. These factors demonstrate the importance of carefully considering the trade-offs between model performance and complexity when approaching classification tasks. In particular, the use of simpler models like Logistic Regression can be beneficial in real-world settings, where interpretability, generalization, and computational efficiency are critical. But in cases where non-linear relationships are present, more complex models like Support Vector Machines may be more appropriate. In any case, it is essential to consider the specific requirements of the task at hand and select the model that best fits these requirements.

\begin{table}[ht]

	\centering
	
	\caption{Comparison of Logistic Regression (LR), Decision Trees (DT), Support Vector Machines (SVM) Models Based on Specific Aspects}

	\renewcommand{\arraystretch}{1.5}

	\begin{tabular}{lllll}
		\toprule
		\textbf{Model} & \textbf{Performance} & \textbf{Classification Capability} & \textbf{Computation Time} & \textbf{Hyperparameter Tuning} \\
		\midrule
		LR      & Superior            & Suitable for binary  & Moderate       & Less complex \\
		                &                     &                     &                &              \\
		\midrule
		DT           & Lower test score mean & Applicable to both  & Least expensive & Moderate complexity \\
		                & Tendency for overfitting & numerical and categorical & & \\
		\midrule
		SVM  & Consistent          & Effective in non-linear & Slowest     & Greater complexity \\
		                &                     & relationships         &                &              \\
		\bottomrule
	\end{tabular}
	\label{table:model_comparison}
\end{table}

\section{Conclusion}

In this study, we have presented a comprehensive comparison of three machine learning techniques, namely Logistic Regression (LR), Decision Trees (DT), and Support Vector Machines (SVM), for the diagnosis of breast cancer using the UCI Machine Learning Repository dataset. Our results demonstrate that the simpler linear model, LR, outperforms the more complex DT and SVM techniques in terms of test score mean, standard deviation, and computational efficiency.

The superior performance of LR can be attributed to several factors. First, its simplicity and interpretability provide a clear understanding of the relationship between input features and the outcome, which is particularly valuable in the context of medical diagnosis. Second, LR's linear decision boundary allows for better generalization and robustness, ensuring reliable performance on new, unseen data. Lastly, the computational efficiency of LR offers advantages in terms of scalability and real-world applicability.

However, it is important to note that Decision Trees and Support Vector Machines showed certain advantages as well. DT models demonstrated faster computation times compared to LR, making them suitable for applications where efficiency is a priority. Additionally, SVM models excel in handling non-linear relationships and high-dimensional spaces with appropriate kernel functions, providing a valuable tool in scenarios where complex decision boundaries are present.

Nevertheless, these benefits of DT and SVM were offset by their respective disadvantages. DT models exhibited a tendency for overfitting, as indicated by the discrepancy between train and test scores. This limitation raises concerns about their generalization capabilities, particularly in the context of medical diagnosis, where reliable predictions on new data are crucial. On the other hand, SVM models required longer computation times compared to LR, which can be a significant drawback in large-scale applications or time-sensitive scenarios.

These considerations highlight the importance of carefully weighing the trade-offs between model performance and complexity when approaching classification tasks. While simpler linear models like LR offer advantages in terms of interpretability, generalization, and computational efficiency, they may not capture the intricacies of non-linear relationships as effectively as more complex techniques like SVM. Therefore, the choice of the model should be guided by the specific problem, the underlying data characteristics, and the desired trade-offs between performance, interpretability, and computational efficiency. In the context of privacy-preservation, simpler models such as logistic regression can often offer advantages over more complex models. One primary reason is their inherent transparency. Logistic regression models are easier to interpret, as they provide clear coefficients for each feature to understand the contribution of that feature to the model's decisions. This interpretability can help identify and mitigate privacy risks, such as the inadvertent inclusion of a sensitive or personally identifiable information (PII) feature. In contrast, complex models like deep neural networks often act as 'black boxes' where the intricate interactions between layers can obscure the contribution of individual features \cite{peake2018explanation}. This opacity can make it difficult in ensuring that these models aren't learning or leaking sensitive information. Moreover, simpler models can perform similar to complex models for many tasks while requiring less data \cite{hastie2009elements}, potentially offering a better privacy-utility trade-off.

Future work could explore the following directions:

\begin{enumerate}
\item \textbf{Feature engineering and selection:} Investigate the impact of feature engineering and selection techniques on the performance of simpler linear models. This could include the use of domain knowledge to create new features, as well as the application of statistical and machine learning methods for feature selection.

\item \textbf{Ensemble methods:} Explore the potential benefits of combining simpler models, such as LR, with more complex techniques, like DT and SVM, through ensemble methods. This could lead to improved performance and robustness by leveraging the strengths of multiple models.

\item \textbf{Model interpretability:} Develop methods to improve the interpretability of more complex models, such as DT and SVM, while maintaining their performance. This could involve the use of visualization techniques, surrogate models, or local explanations to provide better insights into the decision-making process.

\item \textbf{Real-world deployment:} Conduct further studies on the deployment of simpler models like LR in real-world clinical settings, focusing on aspects such as integration with existing workflows, user acceptance, and ethical considerations.

\end{enumerate}

In conclusion, this study emphasizes the power of simplicity in machine learning models for medical diagnosis. By demonstrating the advantages of simpler linear models like Logistic Regression over more complex techniques. We hope that this work will inspire further research in this direction and encourage the use of simpler models in real-world applications.

\newpage



\bibliographystyle{unsrtnat}
\bibliography{references}  

\end{document}